# When normalization hallucinates: unseen risks in AI-powered whole slide image processing

Karel Moens*[a,b], Matthew B. Blaschko [b], Tinne Tuytelaars [b], Bart Diricx[a], Jonas De Vylder[a], Mustafa Yousif[c]

[a] Barco NV, Beneluxpark 21, Kortrijk, Belgium; [b] PSI KU Leuven, Belgium; [c] Michigan Medicine, Ann Arbor, MI USA

**Abstract**

Whole slide image (WSI) normalization remains a vital preprocessing step in computational pathology. Increasingly driven by deep learning, these models learn to approximate data distributions from training examples. This often results in outputs that gravitate toward the average, potentially masking diagnostically important features. More critically, they can introduce hallucinated content, artifacts that appear realistic but are not present in the original tissue, posing a serious threat to downstream analysis. These hallucinations are nearly impossible to detect visually, and current evaluation practices often overlook them. In this work, we demonstrate that the risk of hallucinations is real and underappreciated. While many methods perform adequately on public datasets, we observe a concerning frequency of hallucinations when these same models are retrained and evaluated on real-world clinical data. To address this, we propose a novel image comparison measure designed to automatically detect hallucinations in normalized outputs. Using this measure, we systematically evaluate several well-cited normalization methods retrained on real-world data, revealing significant inconsistencies and failures that are not captured by conventional metrics. Our findings underscore the need for more robust, interpretable normalization techniques and stricter validation protocols in clinical deployment.

## INTRODUCTION

In the field of computational pathology, a significant challenge lies in ensuring that models trained on data from one laboratory can generalize effectively to data from another. This issue arises primarily from the inherent variability in tissue preparation, staining protocols and scanners across different sites, which can lead to considerable differences in image appearance. To address this, researchers have explored two main avenues. The first involves data augmentation techniques applied during the training phase, which artificially increase the diversity of the training data to make the model more robust. The second approach, stain normalization, focuses on standardizing the appearance of images before they are fed into a trained model during inference. While recent studies suggest that sophisticated augmentation strategies can significantly improve generalization, stain normalization methods remain highly prevalent in state-of-the-art pipelines. This is not only true for newly developed AI models but is also a crucial component for integrating with existing third-party systems, which may have already been validated against specific laboratory-defined standards.

Within the realm of stain normalization, two major categories of methods have historically dominated the field. One approach, exemplified by methods like Macenko [1], Reinhard[2] or Vahadane [3], operates on the principle of color deconvolution. These techniques mathematically separate the contributions of different stains, such as hematoxylin and eosin, from the raw image and then reconstruct a normalized version based on a predefined reference. In more recent years, however, a new class of methods based on generative AI has emerged as the leading group. These methods, which initially leveraged Generative Adversarial Networks (GANs) and have more recently included techniques based on Diffusion, have demonstrated a remarkable ability to produce highly realistic and consistent normalized images. A key advantage of these generative methods is their flexibility, e.g. they can easily be adapted to normalize for a wider variety of stains beyond the typical H&E.

Despite their impressive performance and widespread adoption, generative models are known to be susceptible to hallucination, a phenomenon where they introduce spurious features or alter existing ones. This work specifically investigates the intrinsic risk of this on stain normalization. Rather than focusing on the overall performance, which is well-established and a primary reason for their popularity, we turn our attention to the long tail of the data distribution. The goal is to scrutinize how significantly these normalization methods alter the underlying clinical content in outlier cases compared to their raw counterparts. We introduce a novel method for comparing the original input images with their

---

processed normalizations, enabling the systematic identification of potential hallucinations and facilitating a targeted inspection of these outliers. Through this analysis, we have identified a substantial number of disturbing hallucinations, revealing a significant and inherent risk associated with the use of these powerful normalization techniques that warrants careful consideration by the scientific community.

## RELATED WORK

StainGAN [4] introduced a GAN-based approach for whole slide image normalization in histopathology, leveraging generative adversarial networks (GAN) to translate staining styles while preserving tissue structure. Despite its visual effectiveness, GAN-based methods like StainGAN are often challenged by training instability and high sample complexity, which can lead to hallucinated features and unreliable outputs. ContriMix [5] builds on the idea of disentangling content and style through adversarial training. It extracts separate encodings for tissue attributes and staining styles, enabling controlled mixing and augmentation. While this enhances flexibility in data generation, adversarial objectives still inherit the limitations of GANs, including potential instability and feature distortion. In contrast, StainFuser [6] represents a recent shift toward diffusion-based models. It applies noise to input images and reconstructs them in a target staining style using cycle consistency losses to maintain structural fidelity. Although diffusion models like StainFuser offer improved stability over GANs, they risk overwriting subtle but clinically significant features while adding noise, raising concerns about their suitability for diagnostic applications.

## METHODS

In this work, we investigate hallucinations produced by whole slide image normalization methods. To automate the detection of hallucinations and to quantize the severity of hallucinations in test sets, we designed a novel structure discrepancy measure between two images. This structure discrepancy $SD(A, B)$ between grayscale images $A$ and $B$ is defined as follows

$$less\_struct(X, Y) = \frac{S(Y)}{M} \cdot \max\left(0, 1 - \frac{S(X)}{M}\right), \qquad (1)$$

$$SD_{raw}(A, B) = \max(less\_struct(A, B), less\_struct(B, A)) \qquad (2)$$

$$mask(A, B) = \tanh\left(\frac{|A-B|}{P} \pi\right) \qquad (3)$$

$$SD(A, B) = 100 \cdot mask(A, B) \cdot \log(1 + SD_{raw}(A, B)) \qquad (4)$$

where S(X) is the magnitudes of the edges obtained through the application of the Sobel operator, M is a soft threshold on these magnitudes for considering a value as fully textured, and P is a soft threshold on the differences between grayscale pixel values for considering them as different. In equation 4, a mask is applied to ensure that the measure produces low values for similar images, a logarithmic function is used to shorten the tail of the distribution of these values, and a factor of 100 is applied for the readability in visualizations. For smoothness and to capture a wider area of structure, convolutional kernels of 32x32 average out the values of S(A), S(B) and |A-B| in the equations 1-4.

Though still dependent on parameters such as M and P, the qualitative examples in the result section show that this measure is capable of detecting hallucinations produced by normalization methods. In our experiments, it was more effective at this task than measures based on L1, L2 or SSIM between images. A formal comparison and proposal of such measures is outside the scope of this work.

## RESULTS

StainGAN whole slide image normalization models were trained using the code and hyperparameters that were published by the authors. Through analysis with our novel Structure Discrepancy, we then identified hallucinations in an evaluation set. Figure 3 shows the distributions of the discrepancy scores for the training settings listed in Table 1. As a baseline, a StainGAN model was trained and evaluated on CAMELYON16 [7], a popular public dataset. Figure 3 shows that the challenge of training and evaluating on the public CAMELYON16 is indeed misleading compared to the realistic datasets that were used in this study. It is likely that relying on such public datasets will conceal the risk of halucinations. Next, the experiments using subsets of our data seem to indicate that the number of available images for training affects the occurrences of hallucinations at inference. Going from $\frac{1}{16}$ available data to $\frac{1}{8}$ and $\frac{1}{4}$, there is a clear downward shift of the distribution. The gain of adding more training data appears to diminsh from $\frac{1}{4}$ onward. Finally, since whole slide image normalization is a form of domain adaptation to out-of-distribution data, we investigate settings with extreme gaps between the source, target and evaluation data, while still normalizing from one scanner type to another. This lead to the LUNG-TO-SKIN setting, which exemplifies what could happen when a third party model is carelessly applied to data that it was not trained on. Interestingly, the results suggest that a more diverse and challenging training scheme lowers the Structure Discrepency.

Table 1 The training and evaluation settings. CAMELYON16 was added as a baseline and representative of public datasets commonly used in scientific literature. For each dataset, it lists the tissue, scanner and #images, with N a denominator in {16, 8, 4, 2}.

|  | TRAIN SOURCE | TRAIN TARGET | TEST SOURCE |
| --- | --- | --- | --- |
| **CAMELYON16** | Breast, GT450, 6k | Breast, Hamamatsu, 6k | Breast, GT450, 10k |
| **BREAST-TO-BREAST #DATA / N** | Breast, AT2, 65k / N | Breast, GT450, 18k / N | Breast, AT2, 24k |
| **LUNG-TO-SKIN** | Lung, AT2, 50k | Skin, GT450, 18k | Breast, AT2, 24k |

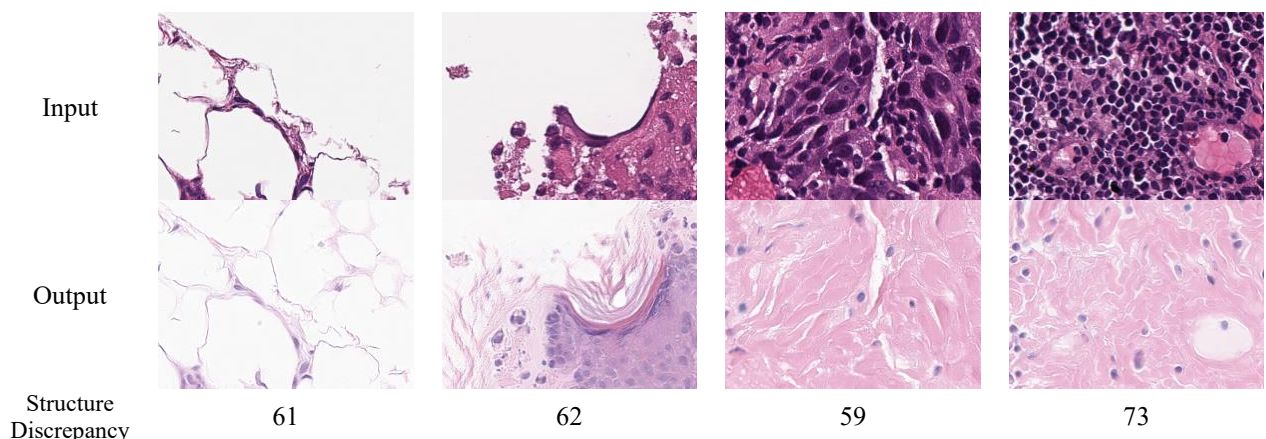

Figure 1 Input and output pairs for two of the trained StainGAN models. The first two columns show hallucinations of the LUNG-TO-SKIN model. The last two columns contain examples of outputs by the BREAST-TO-BREAST model trained on all data.

Besides visual and measurable differences, these hallucinations, identified through our method, can have a clinical impact and pose a real risk. Below we provide brief examples of the clinical impact for the two middle columns of Figure 1. Together with those in Figure 2, these images further highlight the added value of screening AI models with our Structure Discrepancy to test the resilience to hallucinations.

*2nd from the left - Epidermal layer hallucination in AI-generated image.* The original image (top) appears to show breast tissue with no overlying skin or subcutaneous tissue. The AI-generated image (bottom) fabricates additional epidermal strata—including a compact stratum corneum and a prominent stratum granulosum—that are not present in the source. This artificial stratification could be misread as epidermal hyperplasia/hyperkeratosis and materially alter interpretation.

*3rd from the left - Focal necrosis/hemorrhage obscured in AI-generated image.* The original image (top) shows viable tumor with a focal area of hemorrhage/necrosis in the lower-left corner. The AI-generated image (bottom) appears diffusely eosinophilic with loss of hematoxylin signal and nuclear detail, eliminating regional heterogeneity and the discrete lower-left focus. This produces a spurious impression of diffuse nonviability and could markedly distort necrosis estimates and grading.

## CONCLUSION

This investigation into the intrinsic risks of generative AI-based stain normalization methods has revealed a critical and often overlooked vulnerability. While the impressive performance of these techniques on average has led to their widespread adoption, our focus on the long-tail distribution of the data has exposed a significant potential for clinical misrepresentation. This study has demonstrated that despite their ability to produce visually compelling and seemingly normalized images, these models are not immune to hallucination. There is a distinct and important difference between an output that is aesthetically pleasing and one that is a faithful, clinically accurate representation of the original tissue. The proposed comparative methodology has enabled us to systematically identify instances where the normalized output deviates meaningfully from the original clinical content. The disturbing examples we have highlighted underscore that these are not merely aesthetic variations but could potentially impact downstream diagnostic accuracy. The findings from this work do not diminish the value of generative methods in general. They highlight the need for greater transparency and more robust validation protocols that specifically account for the risk of hallucination. Due to page limitations, this discussion has necessarily focused on the general risk of hallucinations in generative stain normalization. The full paper will provide a more elaborate discussion concerning the specific data utilized and the nuances of the training methodologies employed for the generative models. Furthermore, a comprehensive evaluation and detailed discussion of results for other state-of-the-art methods, such as StainFuser and ContriMix, will be presented. Finally, the full paper will delve deeper into the implications of using different training datasets, exploring how variations in their characteristics influences the overall robustness of the normalization process.

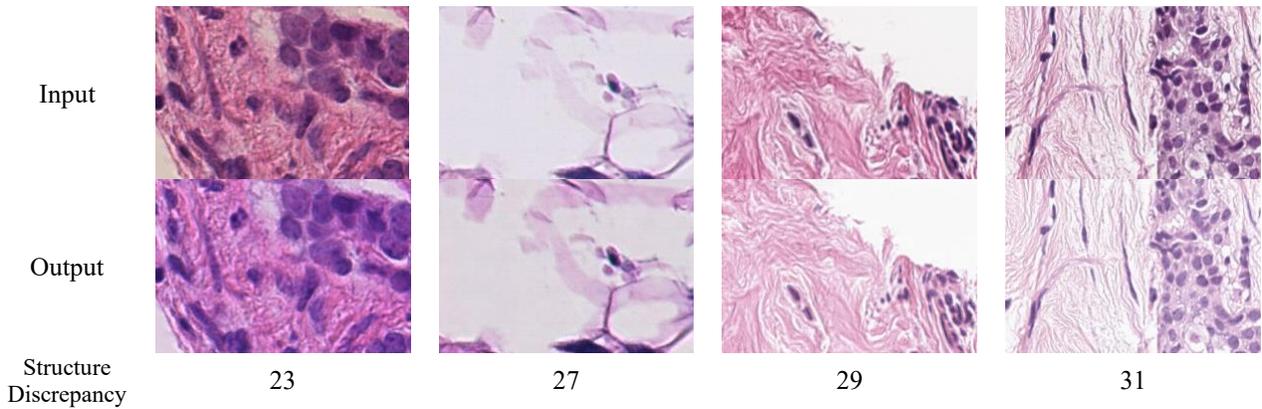

*Figure 2 Input and output pairs for two of the trained StainGAN models. The two pairs on the left are from the evaluation in the CAMELYON16 setting. The two pairs on the right were produced under the BREAST-TO-BREAST setting. These examples illustrate that the code and hyperparameters published by the authors indeed lead to models that function as expected in the majority of cases.*

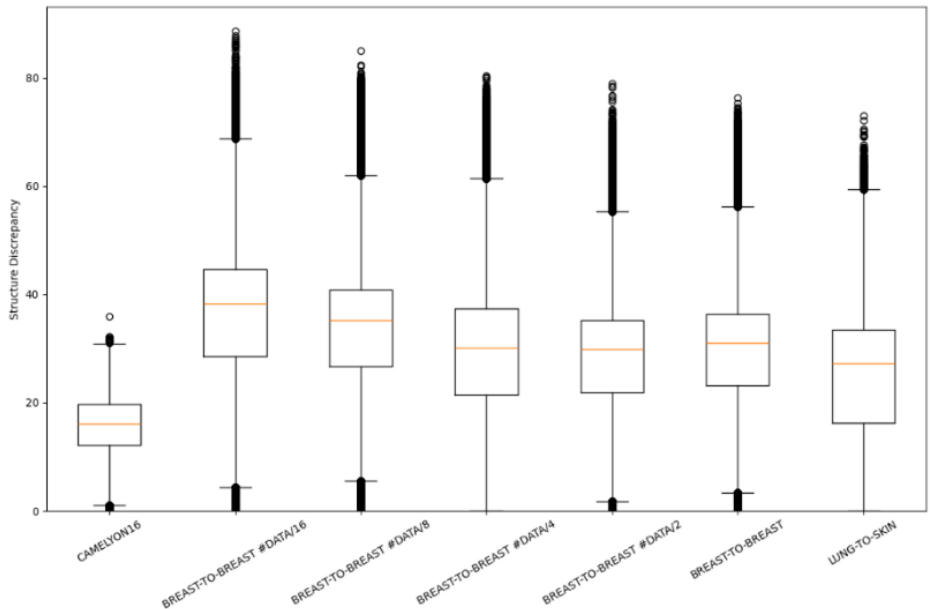

*Figure 3 Boxplots of the distributions of the Structure Discrepancy when evaluating StainGAN models in the settings listed in Table 1.*

# ACKNOWLEDGEMENTS

This research is funded by VLAIO through the LEASTwork project (HBC.2021.0228).